\documentclass{article}
\usepackage{iclr2022_workshop,times}
\usepackage{bbm}
\usepackage{graphicx}
\usepackage{booktabs}
\usepackage{wrapfig}
\usepackage{nicefrac}

\usepackage{hyperref}
\usepackage{url}
\usepackage{amsmath,amsfonts,bm,amsthm,amssymb}

\def\eqref#1{equation~\ref{#1}}

\def\1{\bm{1}}

\def\vh{{\bm{h}}}

\def\mB{{\bm{B}}}
\def\mC{{\bm{C}}}

\def\mH{{\bm{H}}}

\def\mL{{\bm{L}}}

\def\mT{{\bm{T}}}

\def\mW{{\bm{W}}}

\DeclareMathAlphabet{\mathsfit}{\encodingdefault}{\sfdefault}{m}{sl}
\SetMathAlphabet{\mathsfit}{bold}{\encodingdefault}{\sfdefault}{bx}{n}

\def\gN{{\mathcal{N}}}

\def\sR{{\mathbb{R}}}

\newcommand{\darr}{\downarrow}
\newcommand{\uarr}{\uparrow}

\newtheorem{theorem}{Theorem}
\newtheorem{definition}[theorem]{Definition}
\newtheorem{proposition}[theorem]{Proposition}

\title{Simplicial Attention Networks}

\author{Christopher Wei Jin Goh, Cristian Bodnar \& Pietro Li\`o \\
Department of Computer Science\\
University of Cambridge\\
\texttt{\{cwjg4, cb2015, pl219\}@cam.ac.uk}
}

\iclrfinalcopy 
\begin{document}

\maketitle

\begin{abstract}
Graph representation learning methods have mostly been limited to the modelling of node-wise interactions. Recently, there has been an increased interest in understanding how higher-order structures can be utilised to further enhance the learning abilities of graph neural networks (GNNs) in combinatorial spaces. Simplicial Neural Networks (SNNs) naturally model these interactions by performing message passing on simplicial complexes, higher-dimensional generalisations of graphs. Nonetheless, the computations performed by most existent SNNs are strictly tied to the combinatorial structure of the complex. Leveraging the success of attention mechanisms in structured domains, we propose Simplicial Attention Networks (SAT), a new type of simplicial network that dynamically weighs the interactions between neighbouring simplicies and can readily adapt to novel structures. Additionally, we propose a signed attention mechanism that makes SAT orientation equivariant, a desirable property for models operating on (co)chain complexes. We demonstrate that SAT outperforms existent convolutional SNNs and GNNs in two image and trajectory classification tasks.  
\end{abstract}

\section{Introduction \& related work}

\begin{wrapfigure}{r}{0.4\textwidth}
  \begin{center}
      \vspace{-30pt}
    \includegraphics[width=1.0\linewidth]{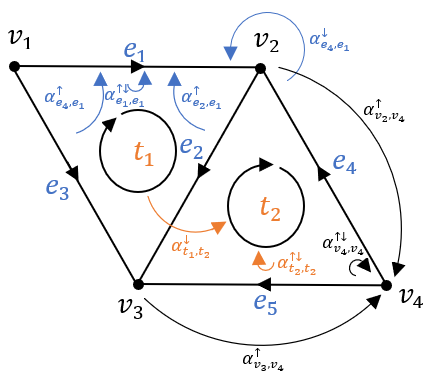}
    \vspace{-20pt}
    \caption{Pictorial diagram of attention coefficients being calculated for node $v_4$, edge $e_1$ and triangle $t_2$}
  \end{center}
    \label{fig:simplicial_attention}
    \vspace{-20pt}
\end{wrapfigure}

Graph Neural Networks (GNNs)
\citep{sperduti1994encoding,goller1996learning,gori2005new,scarselli2008graph,bruna2013spectral,defferrard2016convolutional,kipf2017graph,gilmer2017neural} have become a successful model for statistical problems characterised by an underlying structure. Nonetheless, the simple combinatorial structure of graphs, relying exclusively on dyadic interactions, has recently driven towards extending these approaches to higher-dimensional generalisations such as simplicial~\citep{ebli2020simplicial, bunch2020simplicial, bodnar2021weisfeiler1} and cellular complexes~\citep{bodnar2021weisfeiler2, hajij2020cell} 

Simplicial Nerual Networks (SNNs) have recently been successfully applied to various problems such as missing data imputation \citep{ebli2020simplicial}, graph classification \citep{bunch2020simplicial, bodnar2021weisfeiler1}, molecular property prediction~\citep{bodnar2021weisfeiler2}, edge prediction \cite{chen2021bscnets}, trajectory classification \citep{bodnar2021weisfeiler1} and prediction~\citep{glaze2021principled} as well as homology localisation~\citep{keros2021dist2cycle}. All these approaches can be seen as particular instances of the message passing framework~\citep{bodnar2021weisfeiler1, bodnar2021weisfeiler2}. 

Nonetheless, the computations performed by convolutional approaches are tightly coupled to the combinatorial structure of the complex, which could hinder generalisation to unseen simplicial structures. Therefore, motivated by the recent success of attention-mechanisms in graph representation learning~\citep{velickovic2018graph} and structured domains more generally~\citep{vaswani2017attention}, we propose Simplicial Attention Networks (SAT). Generalising GAT~\citep{velickovic2018graph}, SAT dynamically learns to attend over different neighbouring simplices based on their features. Additionally, we show how SAT can be made orientation equivariant~\citep{glaze2021principled, bodnar2021weisfeiler1} via signed attention coefficients, thus making it suitable for oriented simplicial complexes. 

In practice, we consider an image classification task based on superpixel graphs and a trajectory classification benchmark involving oriented simplicial complexes. We evaluate SAT in both settings against several GNN and SNN baselines and demonstrate it outperforms them. Code for model and experiments can be found at \url{https://github.com/ggoh29/Simplicial-neural-network-benchmark}.

\section{Background}

\paragraph{Simplicial Complexes} Simplicial complexes are a class of topological spaces that are made of nicely glued simplices of various dimensions. Given a set of vertices $V$, a k-simplex is an unordered subset $\{v_0, v_1, \dots , v_k\}$ where $v_i \in V$ and $v_i \neq v_j$ for all $i \neq j$. For a k-simplex $\sigma$ = $\{v_0, v_1, \dots , v_k\}$, we say its faces are all the (k-1)-simplices that are also subsets of $\sigma$, while its cofaces are all (k+1)-simplices that have $\sigma$ as a face. A simplex can also have an orientation, denoted by $[v_0, v_1, \dots , v_k]$, where there is a chosen orientation for its vertices. Two orientations are considered equivalent if they differ by an even permutation, or can be expressed by an even number of transpositions. The choice of orientation is arbitrary and it is used only for bookkeeping purposes.

\paragraph{Adjacencies} Similarly to how we consider two nodes to be adjacent if there exists an edge that connects the two of them together, there is a notion of adjacency for simplicial complexes. However, adjacency can exist in two forms. Two k-simplices $\sigma_i$ and $\sigma_j$ are upper adjacent if both are faces of some $(k+1)$-simplex $\tau$. If the complex is oriented, we further say $\sigma_i$ and $\sigma_j$ are similarly oriented with respect to $\tau$ if the orientations of $\sigma_i$ and $\sigma_j$ agree with the ones induced by $\tau$. If not, they are dissimilarly oriented. Similarly, two k-simplices $\sigma_i$ and $\sigma_j$ are lower adjacent if both have a common face. We denote the upper adjacent simplices of $\sigma$ by $\gN^{\uarr}_\sigma$ and the down adjacent simplices by $\gN^{\darr}_\sigma$. For the purposes of this paper, we also assume that $\sigma \in \gN^{\uarr}_\sigma$ and $\sigma \in \gN^{\darr}_\sigma$. Given two adjacent $d$-dim simplices $\sigma, \tau$, denote by $o_{\sigma,\tau} \in \{\pm1\}$ the relative orientation between them with $o_{\sigma,\sigma}=1$ for all $\sigma$. If the complex is not oriented, we assume $o_{\sigma,\tau} = 1$ for all adjacent simplices $\sigma,\tau$. 

\paragraph{Hodge Laplacian} Consider a simplicial complex $K$. Denote by $\mathbf{C}_k$ the vector space with coefficients in $\sR$ having the oriented $k$-simplices of $K$ as its basis. Elements of this vector space are called $k$-chains. Then we can define a boundary operator $\partial_k : \mathbf{C}_k(X) \rightarrow \mathbf{C}_{k-1}(X)$ acting on the basis elements via $\partial_k [v_0, \ldots, v_k] := \sum_i (-1)^i [v_0, \ldots, \hat{v}_i,\ldots, v_k]$, where $\hat{v}_i$ denotes the face obtained by excluding $v_i$. This can be represented as a matrix $\mathbf{B}_k$ where the rows are indexed by (k-1)-simplices and the columns are indexed by k-simplices. 

Based on the boundary matrix and their transpose, we can define the Hodge Laplacian, a linear operator $\mL_k : \mathbf{C}_k(X) \rightarrow \mathbf{C}_{k}(X)$, which is a higher order generalisation of the graph Lapacian~\citep{lim2019hodge}. In matrix form, the k-th Hodge Laplacian is defined as:
\begin{equation}
\label{eq:1}
    \mL_k = \mB_k^\top \mB_k + \mB_{k+1}\mB_{k+1}^\top 
\end{equation}
One point to note is that $\mB_k^\top \mB_k$ correspond to the lower adjacencies, whereas $\mB_{k+1}\mB_{k+1}^\top $ correspond to the upper adjacencies. We can refer to them as $\mL_k^{down}$ and $\mL_k^{up}$ respectively. The existent simplicial convolutional networks rely on this Laplacian (and its normalised versions) to weigh the different adjacencies between simplices. Instead, we propose using a learned attention matrix that can easily be used both for $k$-chains and arbitrary signals on simplicial complexes. 

\section{Simplicial Attention Networks}

Let $K$ be a simplicial complex. We will describe our model for an arbitrary dimension of the complex $d \leq \mathrm{dim}(K)$. We compute attention coefficients for the up $\alpha^{\uarr}_{\sigma,\tau}$ and down $\alpha^{\darr}_{\sigma,\tau}$ adjacencies via the following equations:
\begin{align}
    \alpha^{\uarr}_{\sigma,\tau} &= o_{\sigma,\tau}\cdot\mathrm{softmax}_{\tau \in \gN^{\uarr}_\sigma}(a(\mW_1 \vh^k_\sigma, \mW_1 \vh^k_\tau)), \\
    \alpha^{\darr}_{\sigma,\tau} &= o_{\sigma,\tau}\cdot \mathrm{softmax}_{\tau \in \gN^{\darr}_\sigma}(a(\mW_2 \vh^k_\sigma, \mW_2 \vh^k_\tau)),
\end{align}
where $a$ is a function for computing attention coefficients. In an oriented simplicial complex, this effectively becomes a form of signed attention. Note that when working at the node-level, where the relative orientation between nodes is trivial and only upper adjacencies are present, one recovers GAT~\cite{velickovic2018graph}. 

Denote by $\mH^d$ the features associated to the $d$-dimensional simplices of $K$. SAT layers are described by the following message passing equation weighting the neighbours by the attention coefficients:
\begin{align}
    \vh^{k+1}_\sigma = \phi \Big(\sum_{\tau \in \gN^{\uarr}_\sigma} \alpha^{\uarr}_{\sigma,\tau} \mW^k_1 \vh^k_\tau, \sum_{\tau \in \gN^{\darr}_\sigma} \alpha^{\darr}_{\sigma,\tau} \mW^k_2 \vh^k_\tau \Big)
\end{align}
Here, $\phi$ is an update function that aggregates the two incoming messages from the lower and upper adjacencies and updates the representation of $\sigma$. More generally, each of these arguments can effectively be augmented with $Z$ attention heads:
\begin{align}
    \vh^{k+1}_\sigma = {\big\Vert}_{z \leq Z} \phi \Big( \sum_{\tau \in \gN^{\uarr}_\sigma} \alpha^{\uarr, z}_{\sigma,\tau} \mW^k_1 \vh^k_\tau, \sum_{\tau \in \gN^{\darr}_\sigma} \alpha^{\darr, z}_{\sigma,\tau} \mW^k_2 \vh^k_\tau \Big)
\end{align}
In oriented simplicial complexes, the choice of orientation is arbitrary and we would like the model to be aware of this symmetry. Mathematically, we would like SAT to be \emph{orientation equivariant}~\citep{bodnar2021weisfeiler1, glaze2021principled}. 

\begin{definition}[\citet{bodnar2021weisfeiler1}]
Let $K$ be a simplicial complex described by boundary matrices $\{\mB_i\}$. A function $f: \mC_k \to \mC_k$ is orientation equivariant if for any diagonal matrix $\mT$ with $\pm1$ on the diagonal, $f(\mT \mH^k, \mB_k \mT, \mT\mB_{k+1}) = \mT f(\mH^k, \mB_k, \mB_{k+1})$.
\end{definition}

We will now show that under certain constraints on the functions $a$ and $\phi$ the model is orientation equivariant.  

\begin{proposition}
If the function $a$ is even in both of its arguments and $\phi$ is odd in both of its arguments, SAT is orientation equivariant. 
\end{proposition}

\begin{proof}[Proof Sketch.]
At the message passing level, the equivariance equation is respected if and only if for any simplex $\sigma$ the local aggregation is invariant to the changes in orientation of the neighbour simplices and if $\sigma$ changes its orientation, the output features $\vh_\sigma^{k+1}$ pick up a minus sign (see \citet{bodnar2021weisfeiler1} for details). 

Because $a$ is an even function, the product $\alpha_{\sigma,\tau} \vh_\tau$ is invariant with respect to changes in the orientation of $\tau$ because the change in sign for $\alpha_{\sigma,\tau}$ and $\vh_\tau$ cancel each other. Therefore, the aggregation is invariant with respect to changes in the orientations of the neighbours. Furthermore, since the function $\phi$ is odd and $\alpha_{\sigma, \sigma} \geq 0$ for any orientation of $\sigma$, it follows that the local aggregation picks up a minus sign when the orientation of $\sigma$ is flipped. 
\end{proof}

\section{Results}

\subsection{Superpixel graphs} 

\paragraph{Dataset} We use the classification of superpixel graphs as our first benchmark to demonstrate the capabilities of SAT. A superpixel graph is the graph representation of an image in which pixels are grouped into nodes representing perceptually meaningful regions, such as a region of similar intensity \citep{SLIC6205760}. By representing the image as a graph, it is possible to change the task of image classification to graph classification. This application of using graph neural networks on superpixel graphs for the task of image classification was first done by \citet{monti} and has since been a popular framework for testing graph neural networks. Here, we use images from the Modified National Institute of Standards and Technology database (MNIST) \citep{mnist} containing handwritten digits from 0 to 9.

\begin{figure}[htb!]
    \centering
    \includegraphics[width=\textwidth]{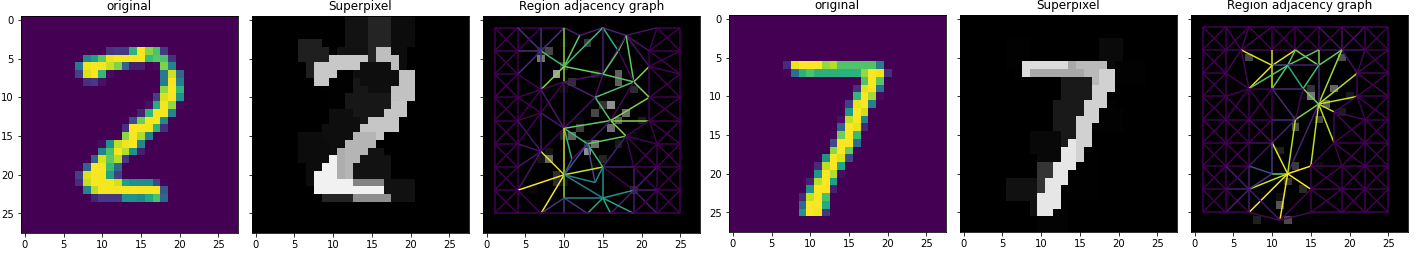}
    \caption{The stages from image to graph for two different MNIST images}
    \label{fig:superpixel_mnist}
\end{figure}
To construct a superpixel dataset from MNIST, we use the Simple Linear Iterative Clustering (SLIC) \citep{SLIC6205760} algorithm. The superpixel nodes are then connected to directly adjacent nodes by means of a region adjacency graph to get the final graph. The simplicial complex we consider is the two-dimensional clique complex of this graph. Triangles are of particular importance as they tend to encode tightly connected regions constituting a form of higher-order interaction between the superpixels. 

We follow the work of \citet{Long_2021} in setting up the node features and how the GNN layers are used. The node features of node cluster $j$ are $\vh_j = [\nicefrac{1}{n} \sum^{n_j}_i (a_i, b_i, g_i)]$ where the $a_i$ and $b_i$ are the x and y coordinate of pixel i respectively, and $g_i$ is the greyscale value of pixel i. The features for the 1-simplices and 2-simplices will be the concatenation of the node features that make up the simplex in the order given by pixel values.

\paragraph{Architecture} Our layers are arranged in the same architecture proposed by \citet{Long_2021} known as a hierarchical GNN architecture. This was used to avoid the over-smoothing problem \citep{cai2020note, oono2019graph}. This architecture comprises of three GNN/SNN layers. The resulting node features from each layer, also known as the residual, are concatenated, resulting in the penultimate feature vector comprising of features vectors from previous layers. This is then passed through a mean pooling function, before being fed to a multi-layer perceptron with 10 outputs followed by a softmax activation. 

\paragraph{Experimental setup} For the experiment, we set the superpixel algorithm SLIC to generate approximately 75 nodes. For all experiments we set a budget of 100 epochs for training, with a batch size of 32 images, using a split of 55k images for training and 5k images for validation. The model corresponding to the epoch with the best validation performance on the validation set is saved and tested against the MNIST test set of 10k images. For the optimisation, we use the Adam optimiser, a weight decay of $0.0005$, and a learning rate of $0.001$ for all models tested. 

We compare SAT against (GCN)~\citep{kipf2017graph}, GAT~\citep{velickovic2018graph}, SCCONV~\citep{bunch2020simplicial}, and SCN~\citep{ebli2020simplicial}. All models were used in a hierarchical GNN setup explained above with 3 layers each. We set the number of feature channels for each model such that the total number of parameters across all models was roughly 10k. All models use a ReLU activation function except SCN which uses Leaky ReLU as in the original paper. Both GAT and SAT use two attention heads.

\paragraph{Results} We repeat the training procedure described above across five seeds and report the mean accuracy and standard deviation. Additionally, we report the exact number of parameters of each model. As the table shows, SAT outperforms its GNN counterpart (GAT), as well as other simplicial convolutional networks that have been recently proposed. 
 
\begin{table}[htb!]
    \centering
    \begin{tabular}{lcccc}\toprule
              Model & Parameters & Test Accuracy \\ \midrule
    \text{GCN} & 10634 & 63.65 $\pm$ 1.82\\
    \text{GAT} & 9862  & 88.95 $\pm$ 0.99\\
    \text{SCN}& 10612 & 84.16 $\pm$ 1.23\\
    \text{SCCONV}& 10315 & 89.06 $\pm$ 0.47\\
    \text{SAT (Ours)} & 10186 & \textbf{92.99 $\pm$ 0.71}\\\bottomrule
    \end{tabular}
    \caption{Image classification accuracy. SAT outperforms its GNN counterpart (GAT) as well as other recently proposed SNNs.}
    \label{tab:img_cls}
\end{table} 

\subsection{Trajectory Classification}

To demonstrate the superior performance of SAT when operating on $k$-chains as well as to display the benefits of orientation equivariance in this setting, we now consider a trajectory classification task~\citep{schaub2020random, bodnar2021weisfeiler1} involving oriented simplicial complexes. 

\begin{wrapfigure}{r}{0.4\textwidth}
    \centering
    \vspace{-18pt}
    \includegraphics[width = 4cm]{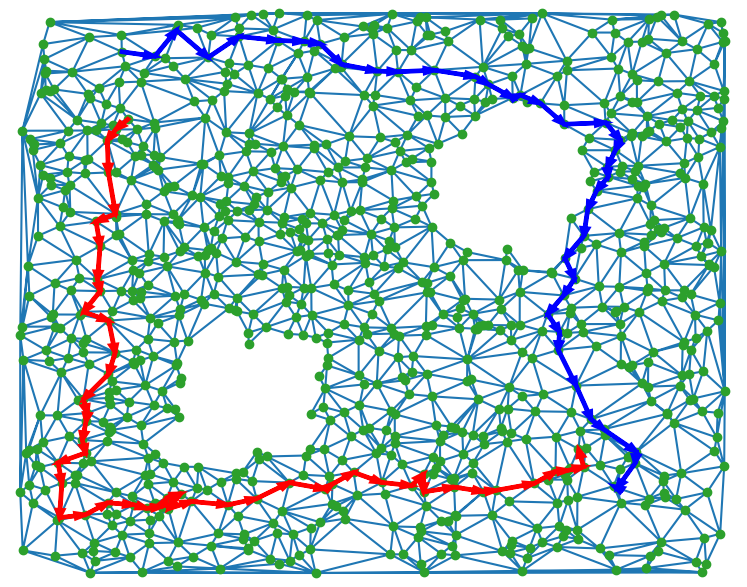}
    \caption{Example of the two different types of trajectories from the trajectory classification dataset.}
    \label{fig:flow}
\end{wrapfigure}

\paragraph{Dataset} The trajectory classification dataset is a synthetic dataset constructed by sampling 1000 points within a unit square followed by a Delauney triangulation. Two holes within the graph are created by removing certain nodes and the edges connected to these nodes. Trajectories are generated by randomly sampling a starting point from the top-left corner and an endpoint from the bottom right corner. Two different types of such trajectories are generated. One class of trajectories traverses via the top-right corner, while the second class traverses via the bottom-left corner. We randomly generate 1000 train and 200 test trajectories in this manner. In order for the test to be challenging for non-orientation invariant models, all trajectories from the training dataset use the same orientation for the edges, while trajectories from the test dataset use random orientations. Two trajectories from this dataset, belonging to two different classes, can be seen in \autoref{fig:flow}.

\paragraph{Experimental setup} All models use the same hyperparameters. We use 4 layers, with the residual size set to 32 for all layers. Following the 4 layers, we take the absolute of the output and do a mean-pooling to make it orientation invariant. We then pass this through two multi-layer perceptron layers before finally taking the softmax of the output. For all experiments, we set a budget of 100 epochs for training, with a batch size of 4. Since there is no validation set, the model corresponding to the epoch with the best performance on the training set is saved and used for testing. For the optimisation, we once again use the Adam optimiser, a weight decay of $0.0005$, and a learning rate of $0.001$ for all models tested.

\paragraph{Results} We repeat the training procedure described above across five seeds and three different activation functions and report the mean accuracy and standard deviation. Models that use odd activation functions (i.e. the identity of tanh), which makes them orientation equivariant, perform better on the test set. Furthermore, the orientation equivariant SAT outperforms other orientation equivariant SNNs. 

\begin{table}[htb!]
    \centering
    \begin{tabular}{lcccccl}\toprule
    & \multicolumn{3}{c}{activation function}
    \\\cmidrule(lr){2-4}
              Model & Id & ReLU & Tanh \\ \midrule
              \text{SCN}& 53.10 $\pm$ 2.27 & 49.70 $\pm$ 2.77 & 52.80 $\pm$ 3.11\\
    \text{SCCONV}& 62.80 $\pm$ 3.11 & 50.80 $\pm$ 1.63 & 62.30 $\pm$ 3.97\\
    \text{SAT (Ours)} & \textbf{92.90 $\pm$ 2.22} & 49.70 $\pm$ 0.60 & \textbf{93.80 $\pm$ 1.33}\\\bottomrule
    \end{tabular}
    \caption{Trajectory classification accuracy. SAT outperforms other recently proposed SNNs.}
    \label{tab:traj_cls}
\end{table} 

\section{Conclusion}

In this paper, we introduced Simplicial Attention Networks (SAT)\footnote{We note that a similar model was proposed concurrently by \cite{simpattentionneuralnet}}, a novel SNN that is able to incorporate attention-mechanisms in order to assign different importance weights to neighbouring simplices as well generalise to unseen simplicial structures. For applications involving orientated simplices, we also illustrated how SAT can be made orientation equivariant via signed attention coefficients. Empirically, we showed SAT outperforms other GNNs and SNNs at the tasks of classifying superpixel images generated from the MNIST dataset and classifying trajectories represented as $1$-chains on a simplicial complex. 

\newpage
\bibliographystyle{iclr2022_workshop}
\bibliography{iclr2022_workshop}

\end{document}